\documentclass[journal]{IEEEtran}
\IEEEoverridecommandlockouts

\usepackage{amsmath}
\usepackage{epsfig} 
\usepackage{amssymb}  
\usepackage{amsthm}  
\usepackage{cite}
\usepackage{bbold}

\usepackage{algorithm} 
\usepackage{algpseudocode} 

\usepackage{epsfig}
\usepackage[caption=false,listofformat=subsimple,labelformat=simple]{subfig}

\usepackage{multirow}
\usepackage{cellspace}
\title{
Reinforcement Learning of Display Transfer Robots \\in Glass Flow Control Systems: \\A Physical Simulation-Based Approach
}

\author{Hwajong Lee, Chan Kim, and Seong-Woo Kim,~\IEEEmembership{Member,~IEEE}

\thanks{Hwajong Lee is with the Graduate School of Engineering Practice, Seoul National University, Seoul, Republic of Korea, 08826, ewioty@snu.ac.kr}
\thanks{Chan Kim is with the Electrical and Computer Engineering, Seoul National University, Seoul, Republic of Korea, 08826, chan\_kim@snu.ac.kr}
\thanks{Seong-Woo Kim is with the Graduate School of Engineering Practice, Seoul National University, Seoul, Republic of Korea, 08826, snwoo@snu.ac.kr}
}

\begin{document}
\maketitle
\thispagestyle{empty}
\pagestyle{empty}

\begin{abstract}
A flow control system is a critical concept for increasing the production capacity of manufacturing systems. To solve the scheduling optimization problem related to the flow control with the aim of improving productivity, existing methods depend on a heuristic design by domain human experts. Therefore, the methods require correction, monitoring, and verification by using real equipment. As system designs increase in complexity, the monitoring time increases, which decreases the probability of arriving at the optimal design. As an alternative approach to the heuristic design of flow control systems, the use of deep reinforcement learning to solve the scheduling optimization problem has been considered. Although the existing research on reinforcement learning has yielded excellent performance in some areas, the applicability of the results to actual FAB such as display and semiconductor manufacturing processes is not evident so far. To this end, we propose a method to implement a physical simulation environment and devise a feasible flow control system design using a transfer robot in display manufacturing through reinforcement learning. We present a model and parameter setting to build a virtual environment for different display transfer robots, and training methods of reinforcement learning on the environment to obtain an optimal scheduling of glass flow control systems. Its feasibility was verified by using different types of robots used in the actual process.
\end{abstract}

\def\abstractname{Note to Practitioners}
\begin{abstract}
This paper is motivated by emerging demands to respond to the various needs of customers by quickly finding optimal manufacturing recipes in the display industry. Although numerical analysis and optimization methods have been proposed for flow control, they assume an ideal process, which makes the results impractical to actual processes. To tackle this problem, we propose a method that builds the physical simulation that reflects physical parameters of display manufacturing processes, and then trains a robot on the simulator. The technical details and considerations are provided, which includes reward design, sensor installation and applicability to actual processes.
\end{abstract}


\section{Introduction}
\begin{figure}[t]
    \centering
    \framebox{\parbox{0.5\textwidth}{\includegraphics[width=0.485\textwidth]{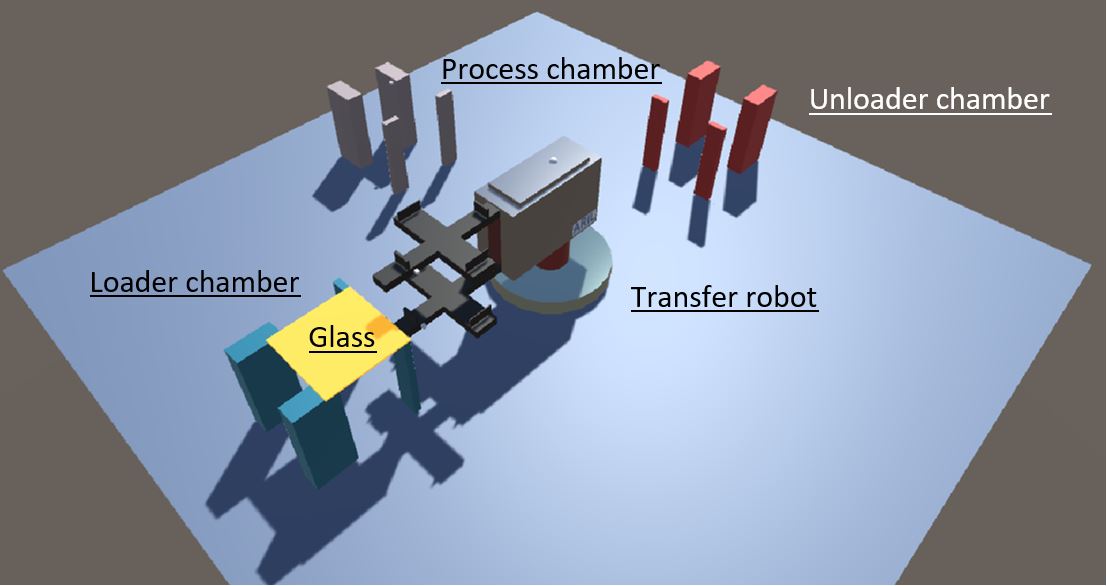}}}
    \caption{Simulation environment configuration.}
    \label{fig:simulation_cofig}
\end{figure}

The global display market continues to expand owing to the proliferation of smartphones, TVs, smartwatches, tablets, and laptops. While fulfilling various market demands and conditions, the display industry is focusing on improving productivity. However, the cost of building a new manufacturing line is astronomical. Manufacturers have strived to maximize the productivity of existing mass production lines. Such efforts have yielded the flow control system, which is closely related to productivity. For example, in display manufacturing fabrication units (FAB), facilities for various purposes such as process, inspection, and measurement are arranged, and glass is conveyed between these chambers according to predetermined conditions, as shown in Fig. \ref{fig:simulation_cofig}. Modern display production involves numerous processes, which increases the complexity of the flow control system.

In a display manufacturing line, as the number of production models increases owing to the high volatility of market demand, the production target of each model changes frequently. In addition, failure and improvement evaluation of equipment for productivity improvement occur frequently during production. These failures and improvement evaluations lower line productivity. Therefore, a method to derive an appropriate flow control system is urgently required for a given production line without stopping production.

\begin{figure*}[t]
    \centering
    \framebox{\parbox{1\textwidth}{\includegraphics[width=1\textwidth]{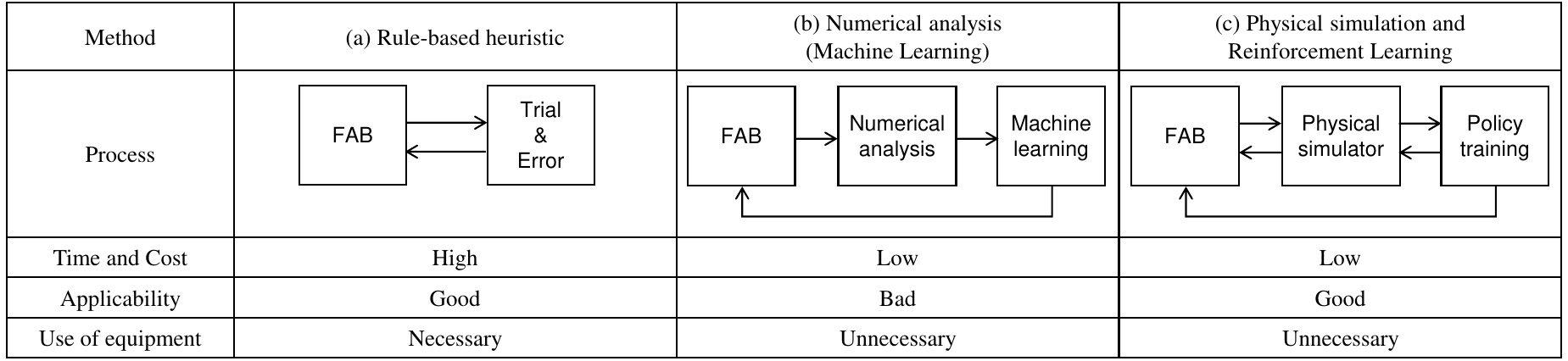}}}
    \caption{Characteristics of existing methods (a) and (b), and proposed method (c).}
    \label{fig:summarytable}
\end{figure*}

The flow control system can be modeled as a flexible jobshop scheduling problem, and various studies have been conducted to solve this problem \cite{defersha2010parallel, qu2005metaheuristic}. Owing to the development of deep reinforcement learning, researchers have been increasingly applying it to solve scheduling problems \cite{waschneck2018optimization, waschneck2018deep, qu2016optimized, arviv2016collaborative}. Most of the studies have focused on the arrangement of the ideal process sequence without using environmental information of the actual manufacturing process. However, in an actual manufacturing line, multiple variables are in play, such as the position of the process chambers, movement speed of the transfer robot, weight and size of the production model, and chamber conditions. The time interval between processes changes continuously depending on these variables. In the design of a flow control system, the environment should reflect real-world variables. 


In this paper, we propose a method that can consider the variables related to the physical elements of an actual manufacturing process. For the sake of this, we build a virtual environment that reflects the physical and process parameters of a manufacturing environment, as shown in Fig. \ref{fig:simulation_cofig}, and then perform reinforcement learning to improve the performance of the flow control system. In the method, we consider the flow control system environment of a FAB unit process, the environment of which is configured using a physical simulator.

At the process entrance, glass is input at regular time intervals, where this glass is transferred to the target point after a set of processes. The transfer robot, an agent trained using reinforcement learning, learns the scheduling problem of flow control system by interacting with the physical simulation environment. We define the state, action, and reward for training the transfer robot. The simulated environment is characterized by physical parameters and process parameters. Based on the virtual FAB environment, policy learning is performed to improve the performance of the flow control system performance for the OLED display manufacturing process. To the best of our knowledge, there is no prior research on the reinforcement learning of glass flow control systems using display transport robots in a digital twin manner.

\indent The main contributions of this paper are as follows:
\begin{itemize}
\item We present a modeling framework and approach to build a virtual FAB environment for optimal scheduling of a display transfer robot in flow control.
\item We present a training method to obtain an optimal control policy on the virtual system using reinforcement learning frameworks for display manufacturing process.
\item We share the practical issues to reduce the gap between actual policy and learned policy from virtual environments, which includes how to design appropriate rewards and metrics, what sensors and where to install. 
\end{itemize}

It is too tedious and tricky to find a working policy of learning-based robot control in a digital twin manner that consider physical parameters. Up on the results and experience presented in this paper, we hope to see the development of diverse learning-based robot control and deploy them to actual manufacturing processes.


\section{Background \& Related Works}
Globally, the mobile and TV markets are transitioning from LCD displays to cutting-edge OLED displays, and the high cost of the equipment required for manufacturing OLED displays imposes a heavy investment burden on manufacturers. Due to the rapid development of display manufacturing technologies, rapidly aging components, and short service lifetime, the initial setup time has become a  critical factor for display equipment that is becoming increasingly complex, automated, and highly functional.

The factors that must be considered in the construction of a new line include the size and weight of glass, time required by the process equipment, required time interval between various pieces of process equipment, need for a glass storage buffer, and type and specification of the transfer robot. It is based on the layout of the previous version, but a new facility layout is required when the considerations change. Significant amounts of review and testing times are needed to fulfill these requirements. Moreover, imbalances in the flow control system after line setup hinder reconfiguration or warrant significant reinvestment. In the display industry, new products are developed depending on the volatility in market demand, and the production requirements of each product change frequently. Accordingly, the setup status of equipment keeps changing. To reflect these changes in the production environment, various pieces of equipment in a line have to be stopped and tested, which reduces line productivity. See the Fig. \ref{fig:summarytable}(a).

\begin{figure*}[t]
    \centering
    \subfloat[Timetable according to existing research.]{
        \framebox{\parbox{1\textwidth}{\includegraphics[width=1\textwidth]{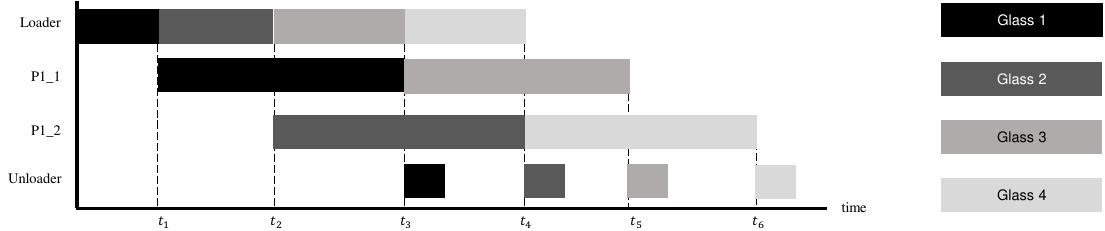}}}
        \label{fig:timetable1}
    }\\
    \subfloat[Timetable of real FAB.]{
        \framebox{\parbox{1\textwidth}{\includegraphics[width=1\textwidth]{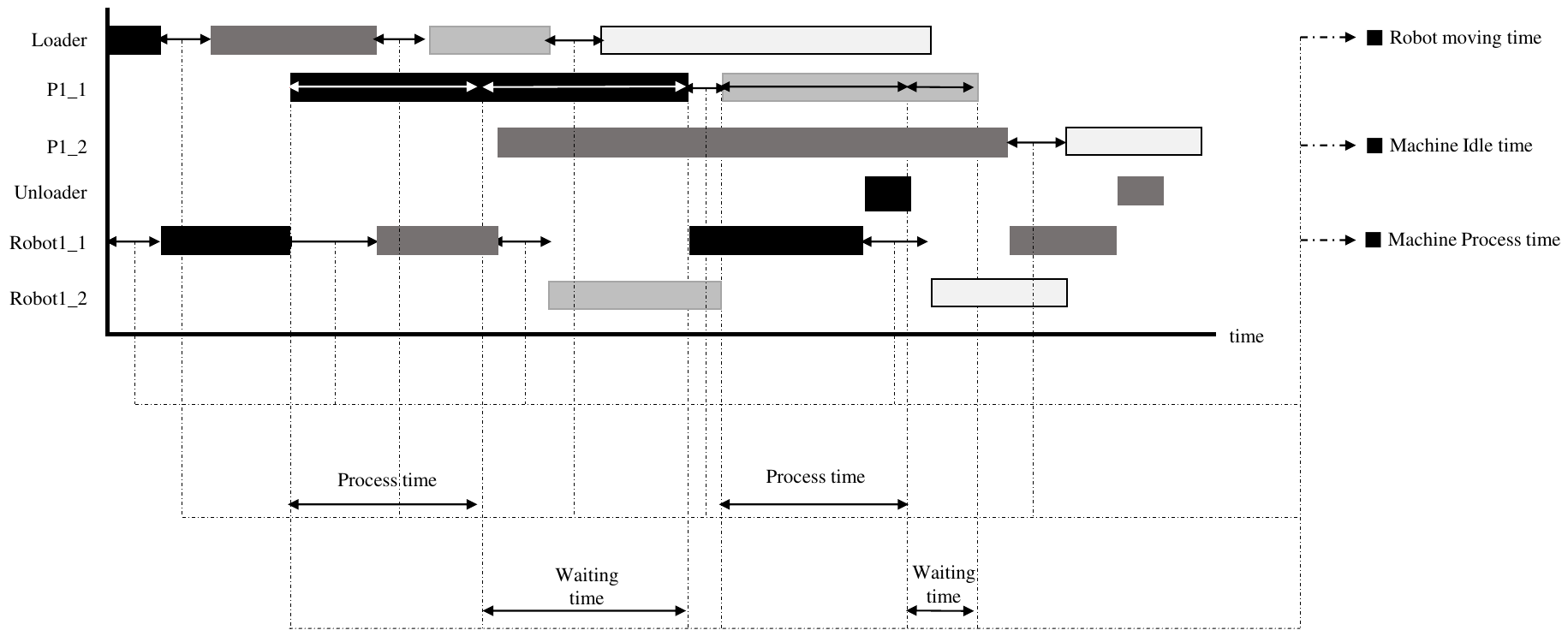}}}
        \label{fig:timetable2}
    }
    \caption{Existing studies that do not consider the important scheduling conditions of actual FAB units.}
\end{figure*}

Previous studies have focused on the optimization of process arrangement by using the methods proposed for process tasks under certain conditions, and the results are presented as numerical data, as represented in Fig. \ref{fig:summarytable}(b). However, the methods proposed therein have clear limitations. For example, if the methods are applied to the two-arms and four chambers layout, the timetable in Fig. \ref{fig:timetable1} is obtained. The vertical axis represents equipment, and the horizontal axis represents time. Although this appears to be an ideal glass transport sequence, its applicability is poor in actual FAB units, because an actual FAB timetable includes robot moving time, machine idle time, and machine process time, as shown in Fig. \ref{fig:timetable2}. The previous studies do not fit with the consideration of the important scheduling conditions of actual FAB units. 

As a second limitation, various process parameters in the real environment are not considered, including glass input interval, size, weight, number of chambers, chamber arrangement, speed of transfer robot, number of robot arms, and different process times for individual processes. These process parameters impart great variability to the flow control system. In other words, although the existing numerical studies for solving the problem associated with the rule-based heuristic method may be numerically excellent, they are limited in terms of their applicability in actual FAB units. Consequently, there are very few cases in which these methods have been applied to an actual process such as display manufacturing process.

\begin{figure}
    \framebox{\parbox{0.5\textwidth}{\includegraphics[width=0.5\textwidth]{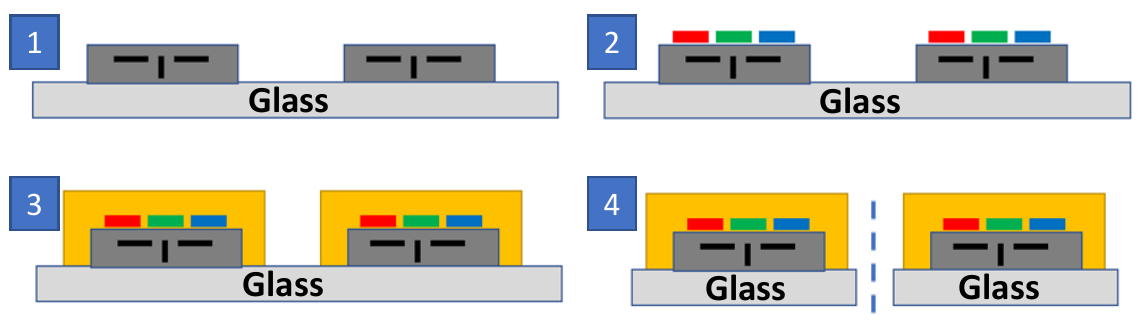}}}
    \caption{OLED manufacturing process: 1. Backplane process, 2. Evaporation process, 3. Encapsulation process, 4. Cell and module process.}
    \label{fig:OLED}
\end{figure}

\subsection{OLED display process}
Fig \ref{fig:OLED} shows the OLED display manufacturing process, which consists of four major steps.
\begin{itemize}
\item \emph{Backplane process}: This is the process of stacking a thin film transistor (TFT) layer that switches on or off power to each pixel and controls brightness by supplying and controlling current.
\item \emph{Evaporation process}: This is the process of depositing an organic fluorescent material on the TFT layer and realizing various colors by mixing RGB.
\item \emph{Encapsulation process}: This process covers the RGB organic fluorescent material with a thin film and protects the internal organic materials and electrodes from oxygen and moisture.
\item \emph{Cell and Module process}: This process cuts the OLED panel into cells and modularizes it.
\end{itemize}
\indent \;\; Each unit process consists of several process chambers and a transfer robot. Reducing the movement lines between each process is beneficial for increasing production and securing quality. Even within individual processes, the scheduling of glass transfer between process chambers is extremely important. Multi-chamber configurations, such as cluster-type arrangements, are commonly used. A lack of proper scheduling leads to defects or reduced production. Currently, most display manufacturing sites use the heuristic rule-based return logic. Each process chamber sends an individual in-out signal, and the transfer robot executes the fastest signal that satisfies the process sequence and conditions.

\subsection{Reinforcement learning for flow control system}
Many researchers have attempted to study reinforcement learning with the aim of optimizing scheduling. The behavior of selecting the best rule among various available rules through reinforcement learning has been determined. Specifically, in the manufacturing industry, the scheduling problem has been studied by applying reinforcement learning to in-line equipment connected in a single facility \cite{yuan2013dynamic}. The multi-agent reinforcement learning method has been studied to derive the optimal schedule of a process system composed of various machines for manufacturing diverse types of products \cite{qu2016optimized}. 

Also, studies have been conducted to solve the jobshop scheduling problem. In these studies, the reinforcement learning agent learns the behavior of selecting the priority rule in a real-time jobshop system. Through this, Aydin \textit{et al.} confirmed the possibility of using a reinforcement learning agent in a dynamic scheduling system \cite{aydin2000dynamic}. Gabel \textit{et al.} proposed a method to solve the scheduling problem with a distributed reinforcement learning agent by using a policy gradient (PG)-based model \cite{gabel2012distributed}. In the context of dual-armed control for multiple chambers, the problem setting of Hong and Lee \cite{hong2018multi} in semiconductor manufacturing is similar to the target problem of this paper. However, including the Ref. \cite{hong2018multi}, all the above-mentioned studies have assumed ideal processes for numerical analysis without using environmental information of the actual manufacturing process.

\subsection{Virtual environments in manufacturing}

In simulation methods, it is essential to implement a sophisticated virtual environment that minimizes the gap between actual target environment and virtual environment to obtain optimal policy, which can be categorized as a digital twin system \cite{fuller2020digital, shangguan2019hierarchical}. Most of the current research platforms for realizing optimal control are based on game engines or video games, such as Atari, Minecraft, and Doom \cite{mnih2015human, jin2018regret, guss2019minerl}. Unlike the game engines, it is difficult to apply reinforcement learning to real-world problems due to such unavailability of the corresponding virtual environments.



In previous studies based on simulation techniques, Dayhoff and Atherton  introduced a simulation model of the FAB process and used various simulation models to analyze system performance\cite{dayhoff1986signature}. In another study \cite{atherton1986signature}, the same authors used signal analysis techniques to describe the characteristics of semiconductor wafer processing and test the effects of various work assignment rules. To determine the yield of wafers entering the FAB process, the performance of the system was tested using simulation model data. 

Sakr \textit{et al.} proposed a reinforcement-learning-based application for dispatch and resource allocation in semiconductor manufacturing \cite{sakr2021simulation}. This application is based on a discrete event simulation model of a real semiconductor manufacturing system. It simulates various aspects of processing that are commonly present in complex systems. The agent in the model uses Deep-Q-Network and simultaneously learns through model execution. Hildebrand \textit{et al.} presented a learning-based method for a robot batching process by building virtual environments and reinforcement learning \cite{hildebrand2020deep}.

As far as the authors know, there is no prior research on the reinforcement learning of glass flow control systems using display transport robots in a digital twin manner, as proposed in Fig. \ref{fig:summarytable}(c).

\section{Physical simulation application and learning}

It is necessary to design an optimal layout without actual equipment and appropriately respond to changes in the production environment without stopping the production equipment. Therefore, we propose a method to derive an optimal layout and minimize facility usage time by implementing a virtual process environment through physical simulation and examining various aspects of logistics flow, process balance, and production management in advance. We use a basic display FAB unit process for this purpose. The selected unit process is composed of four facilities (Fig. \ref{fig:simulation_cofig})

\begin{figure}[t]
    \centering
    \framebox{\parbox{0.4\textwidth}{\includegraphics[width=0.4\textwidth]{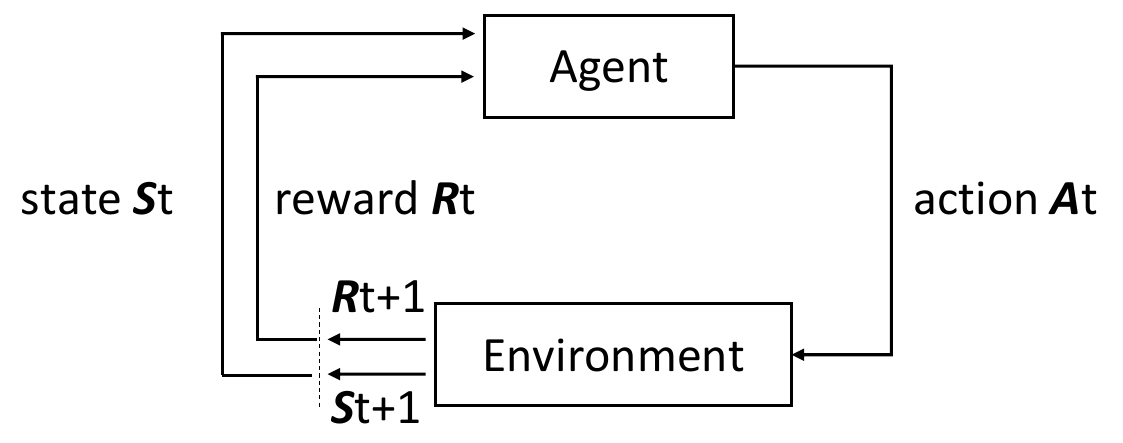}}}
    \caption{Basic structure of Reinforcement learning.}
    \label{fig:RL}
\end{figure}

\begin{figure*}[t]
    \centering
    \framebox{\parbox{1\textwidth}{\includegraphics[width=1\textwidth]{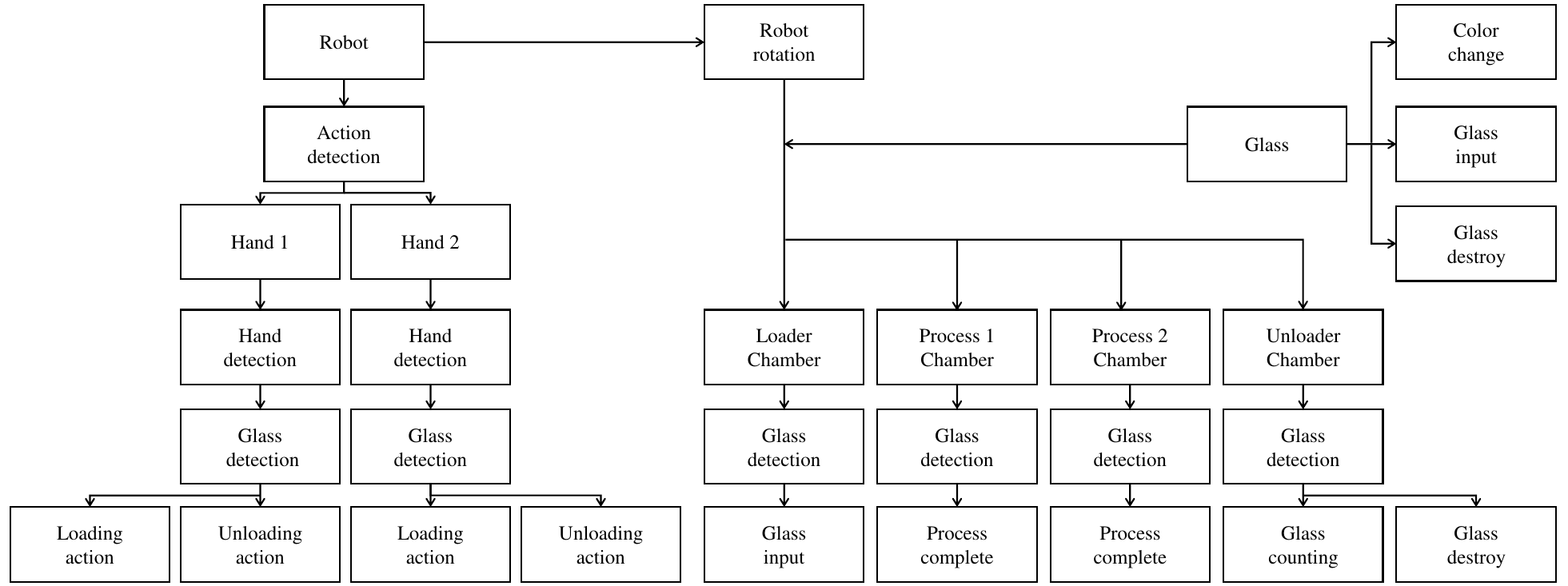}}}
    \caption{Software Configuration.}
    \label{fig:codemap}
\end{figure*}

\begin{itemize}
\item \emph{Loader chamber}: The loader chamber is the first starting point of production rate management. It can be the first inlet to the line or the glass input chamber of the unit process.
\item \emph{Transfer robot}: This robot is used for moving glass between chambers. It has more than one arm, and it acts based on the decisions made by the flow control system.
\item \emph{Process chamber}: In this chamber, various technologies are used for completing each process, and the process time depends on the purpose. Multiple facilities are designed in various layouts to increase production.
\item \emph{Unloader chamber}: This is the final target point for the completion of glass processing, meaning that it is the last facility for production rate control.
\end{itemize}
The following equation is used to compute the time required for glass to enter the loader chamber and exit the unloader:
\begin{equation}
    \begin{aligned}
    Process\;\;tact = +t_{R_{a}}+t_{get}+t_{R_{b}}+t_{put}+t_{P}\\+t_{w}+t_{get}+t_{R_{c}}+t_{put}+t_{U},
    \end{aligned}
\end{equation}
where
\begin{itemize}
\item \emph{$t_{L}$}: Loading time;
\item \emph{$t_{U}$}: Unloading time;
\item \emph{$t_{R}$}: Rotation time;
\item \emph{$t_{get}$}: Glass get time;
\item \emph{$t_{put}$}: Glass put time;
\item \emph{$t_{p}$}: Process time; and 
\item \emph{$t_{w}$}: Waiting time.
\end{itemize}

If we generalize this expression to the case of K process chambers, we have
\begin{equation}
    \begin{aligned}
    t_{L}+\sum_{1}^{K}t_{R_{(n)}}+(K+1)(t_{get}+t_{put})\\+\sum_{1}^{K}t_{P_{(n)}}+\sum_{1}^{K}t_{w_{(n)}}+t_{U}.
    \end{aligned}
\end{equation}

This equation is divided into the following process parameters items: $t_{L}+(K+1)(t_{get}+t_{put})+\sum_{1}^{K}t_{P_{(n)}}+t_{U}$ and the  $\Delta t$ item $\sum_{1}^{K}t_{R_{(n)}}+\sum_{1}^{K}t_{w_{(n)}}$. The latter one is changed according to the movement of robot transport. In a physical simulation, various process parameters can be applied, and the term $\Delta t$ determined by the robot’s transport can be implemented. Because the use of actual equipment is unnecessary, there is no risk of damage to equipment and glass.

\begin{algorithm}[t]
	\caption{Proximal Policy Optimization \cite{schulman2017proximal}} 
	\begin{algorithmic}[1]
		\For {$iteration=1,2,\ldots$}
			\For {$actor=1,2,\ldots,N$}
				\State Run policy $\pi_{\theta_{old}}$ in environment for $T$ time steps
				\State Compute advantage estimates $\hat{A}_{1},\ldots,\hat{A}_{T}$
			\EndFor
			\State Optimize surrogate $L$ wrt. $\theta$, with $K$ epochs and minibatch size $M\leq NT$
			\State $\theta_{old}\leftarrow\theta$
		\EndFor
	\end{algorithmic} 
\end{algorithm}

The detailed system configuration is as follows. The first step is the simulation of a real FAB. The simulator used here is the Unity, and the equipment consists of a transfer robot, chamber, and glass. As shown in Fig. \ref{fig:simulation_cofig}, the loader chamber, unloader chamber, and process chamber are arranged in consideration of the transfer robot. The next task is software configuration. We used \(\textrm{C}\textrm{\#}\) and created a sensor that detects actions based on the robot, robot arm, chamber, and glass. A glass detection sensor is added because the action changes depending on the presence of glass. We implemented the actions that are directly executed by each hardware component. The software configuration is the same as that in  Fig. \ref{fig:codemap}.

The second step is process parameter optimization. In the simulation environment, it is possible to evaluate various process parameters as in case of the real FAB unit. Representative process parameters include glass input interval, glass size, weight, number of chambers, chamber arrangement, speed of transfer robot, number of robot arms, and process time. The parameter data obtained through evaluation are updated using the Recipe Management System (RMS) and Equipment Constants Management (ECM) system.

The third step is the training phase. Reinforcement learning is performed to learn the simulator and improve the performance of the flow control policy. Reinforcement learning is a branch of machine learning, which consists of an agent, an environment, state, a reward, and an action. At each point in time, the agent observes the state of the environment and receives a reward. The goal of the agent is to maximize its reward \cite{sutton2018reinforcement}. 

Fig. \ref{fig:RL} shows the basic process of reinforcement learning. Markov decision process (MDP) is a mathematical concept that enables one to provide theoretically sophisticated explanations of reinforcement learning problems. MDP is composed of $S,A,R,P$, which denote state, action, reward, and transition probability, respectively. As shown in the Fig. \ref{fig:RL}, the agent receives state $S_t$ from the environment. Then, it executes action $A_t$ according to $S_t$. The Environment receives $A_t$ and accordingly gives $S_{t+1}$ and Reward $R_t$ to the Agent.

\begin{figure}[t]
    \centering
    \framebox{\parbox{0.30\textwidth}{\includegraphics[width=0.30\textwidth]{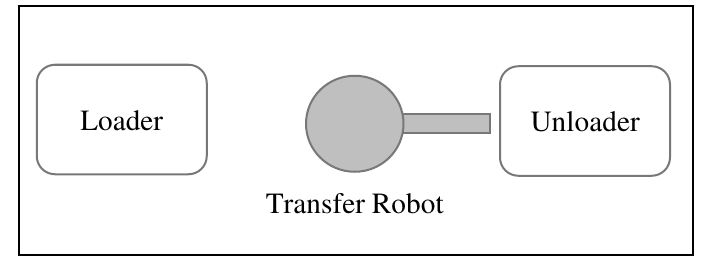}}}
    \caption{A schematic diagram of a transfer robot.}
    \label{fig:basicenv}
\end{figure}

The proximal policy optimization (PPO) algorithm \cite{schulman2017proximal} is used for learning in this study. The PPO algorithm creates a trust region to shield problems against excessive policy updates occurring in the PG, thus facilitating stable neural network updates. Similar to the existing policy-based reinforcement learning, the policy neural network $\pi _{\theta}(a|s)$ interacts with the environment and collects data in memory. In this algorithm, the loss function for neural network update is expressed as Equation \eqref{eqn:0} by using the value multiplied with importance sampling after collecting the reward values for each time step.
\begin{equation}\label{eqn:0}
    \emph L(\theta )=E_{s\sim p^{\pi _{\theta _{dd}}}},_{a\sim {\pi _{\theta _{old}}}}[\frac{\pi _{\theta}(a|s)}{\pi _{\theta_{old}}(a|s)}\hat{A}_{\theta_{old}}(s,a)].
\end{equation}
In PPO, $r_t(\theta)$ is limited to a constant $\epsilon$ (usually 0.2), and the resulting $L$ is expressed as Equation \eqref{eqn:2}.
\begin{equation}\label{eqn:2}
    \emph L^{clip}(\theta)=\hat{E}_t[min(r_t(\theta)\hat{A}_t,clip(r_t(\theta),1-\epsilon ,1+\epsilon )\hat{A}_t],
\end{equation}
where
\begin{equation}\label{eqn:1}
    \emph r_t(\theta)=\frac{\pi _{\theta}(a|s)}{\pi _{\theta_{old}}(a|s)},r_t(\theta_{old})=1.
\end{equation}
Thereafter, similar to a normal neural network update, the data accumulated in memory are divided into batches. The data are updated $K$ times in order by using loss, and the process from the top to the present is repeated continuously, as in Algorithm 1. Because we are using a single robot agent for learning, the number of actors is one.

\begin{figure}[t]
    \centering
    \framebox{\parbox{0.5\textwidth}{\includegraphics[width=0.48\textwidth]{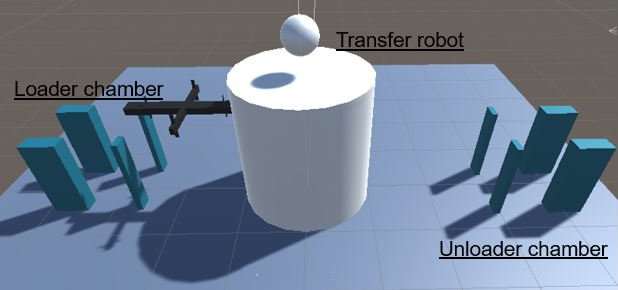}}}
    \caption{Basic experiment environment.}
    \label{fig:basicenv_2}
\end{figure}

The fourth and final step is model creation and application. In this stage, intermediate models are generated continuously during learning. Upon the completion of training, the agent brain file of the final model is created. The file can be applied to the simulation environment or the real FAB environment.

In this section, we have provided a comprehensive methodology. However, the most difficult and trickiest part is to design the reward and to consider the  physical characteristics of actual manufacturing, which will be explained in detail in the following section.

\begin{table}[t]
\caption{Physical parameter setting.}
\label{table:7}
\begin{center}
    \begin{tabular}[\textwidth]{c c || c c}\hline
    Item & Value & Item & Value \\\hline
    Glass scale & x:1 / y:0.01 / z:1 & Mass & 500 \\
    Drag & 0 & Angular drag & 0.05 \\
    Gravity & Use & Dynamic friction  & 0.6 \\
    Static friction & 0.6 & Transfer speed    & 50f  \\
    Process time & 50f & & \\\hline
    \end{tabular}
\end{center}
\end{table}

\section{Experiment Set-up}
The experiment proceeds in two stages, namely basic experiment and extension experiment. The actual FAB is simulated in the basic experiment, and the simulation training is verified through reinforcement learning. A process chamber is added, and an expansion experiment is conducted using a two-arm transfer robot.
\subsection{Basic experiment}
\subsubsection{Simulation configuration}
The simulation setup is composed of a one-arm transfer robot, a loader chamber, and an unloader chamber in Fig. \ref{fig:basicenv} and Fig. \ref{fig:basicenv_2}. To implement conditions similar to those in an actual FAB environment, the following characteristics are imparted to the major facilities.

\begin{itemize}
\item \emph{Loader chamber}: Glass is created at regular time intervals. If there is glass in the chamber, the input is stopped, and the next glass is input after the current glass is out.
\item \emph{Transfer robot}: It is composed of an arm that extends toward the chamber along the x-axis, performs lifting and lowering of the glass along the z-axis, and moves the glass along the theta axis. The corresponding actions are as follows: loading to get the glass, rotation to move the glass to the next destination, and unloading the glass.
\item \emph{Unloader chamber}: This is the goal of the simulation to achieve. When the glass enters the chamber, it disappears, and the number of glasses is counted.
\end{itemize}

\begin{figure}[t]
    \centering
    \framebox{\parbox{0.5\textwidth}{\includegraphics[width=0.5\textwidth]{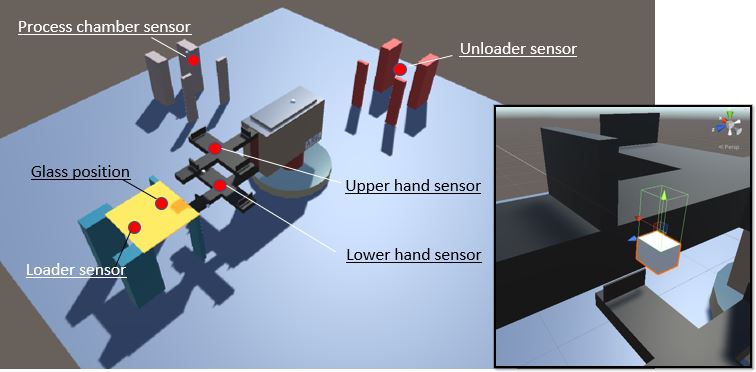}}}
    \caption{Observation of simulation (Object position).}
    \label{fig:position}
\end{figure}

\begin{figure*}[t]
    \centering
    \framebox{\parbox{1\textwidth}{\includegraphics[width=1\textwidth]{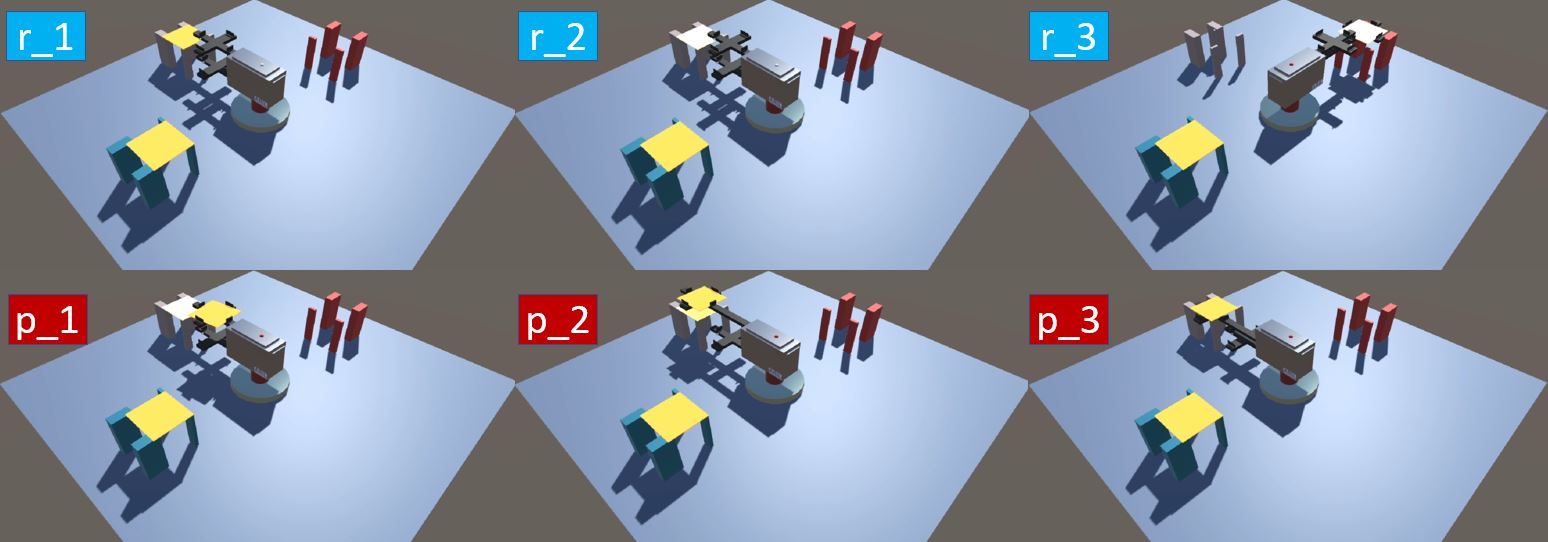}}}
    \caption{Simulation reward (r): for obtaining the reward, the processed glass (white) must be moved to the unloading chamber, (p): As a typical penalty scenario, collision of the robot arm with the glass is depicted.}
    \label{fig:reward}
\end{figure*}

\begin{table}
\caption{Hyper parameters in PPO algorithm (Basic experiment).}
\label{table:k}
\begin{center}
    \begin{tabular}{c c || c c || c c }\hline 
    Parameter & Value & Parameter & Value &  Parameter & Value \\ \hline 
    Batch size & 5120 & $\beta$ & 500 & Buffer size & 102400 \\
    $\epsilon$  & 0.2 & $\gamma$ & 0.99 & $\lambda$ & 0.95 \\
    Learning rate & 3.0e-4 & Max steps & 5.0e7 & Memory size & 256 \\
    \hline
    \end{tabular}
\end{center}
\end{table}

\begin{figure}[t]
    \centering
    \framebox{\parbox{0.38\textwidth}{\includegraphics[width=0.38\textwidth]{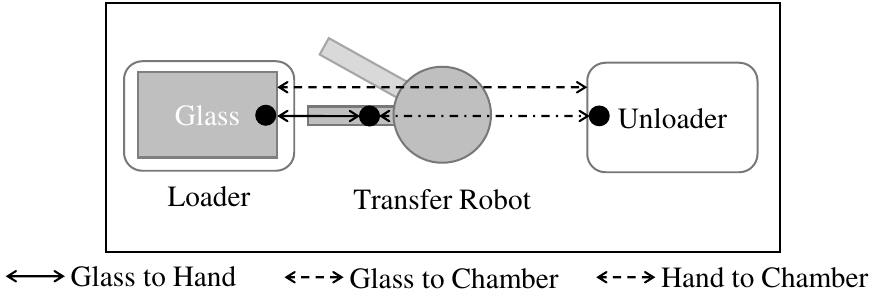}}}
    \caption{Observation of simulation (Object distance).}
    \label{fig:distance}
\end{figure}

The arm of the transfer robot can grab, move, and place only one glass; when the arm collides with the glass, the glass is destroyed and disappears. In the same way as an actual FAB transfer robot, a glass guide pin is installed in the arm. The physical parameter settings are summarized in Table \ref{table:7}.

Loading and unloading of the glass and rotation to the chamber position are possible. An interlock is set such that when one action is executed, other actions are not executed simultaneously.

\subsubsection{Simulation learning through reinforcement learning}

For learning the flow control system, reinforcement learning was performed in a Python environment by using ML agents. The performance of reinforcement learning is affected by the definitions of state, action, and reward components of MDP.

The transfer robot, which acts as an agent, requires state observations for learning. State includes location information of the glass, chamber, and robot, as well as the distance information between these components. Action is a decision that the reinforcement learning agent must select based on the state. The system checks the presence or absence of glass in the process chamber. Action is defined by the combination of the glass movement action of each arm and the rotation toward the chamber. 

The number of transfer robot movement actions is the same as the number of chambers, and the number of arm actions is twice the number of robot arms. Therefore, the total number of actions can be defined as (number of chambers + number of arms$\times$2). The number of actions of an arm is doubled because each arm can perform both loading and unloading operations. Moreover, there is a waiting action, in which the agent does not execute any operation. The movement and the glass loading and unloading operations of an arm cannot occur simultaneously. There are no episode endings for learning in a continuous environment.

Fig. \ref{fig:reward} depicts the scenarios of reward and penalty, whose rewards are defined in Table \ref{table:r}. Reward 1 is incurred when the target position on the glass is reached upon process completion. Penalty 1 is time penalty, Penalty 2 is the penalty incurred when the glass is dropped from the robot arm, Penalty 3 is the penalty incurred due to collision of the robot arm with the glass, and Penalty 4 is incurred when the target position is reached on the incompletely processed glass.

\begin{figure*}[t]
    \centering
    \framebox{\parbox{1\textwidth}{\includegraphics[width=1\textwidth]{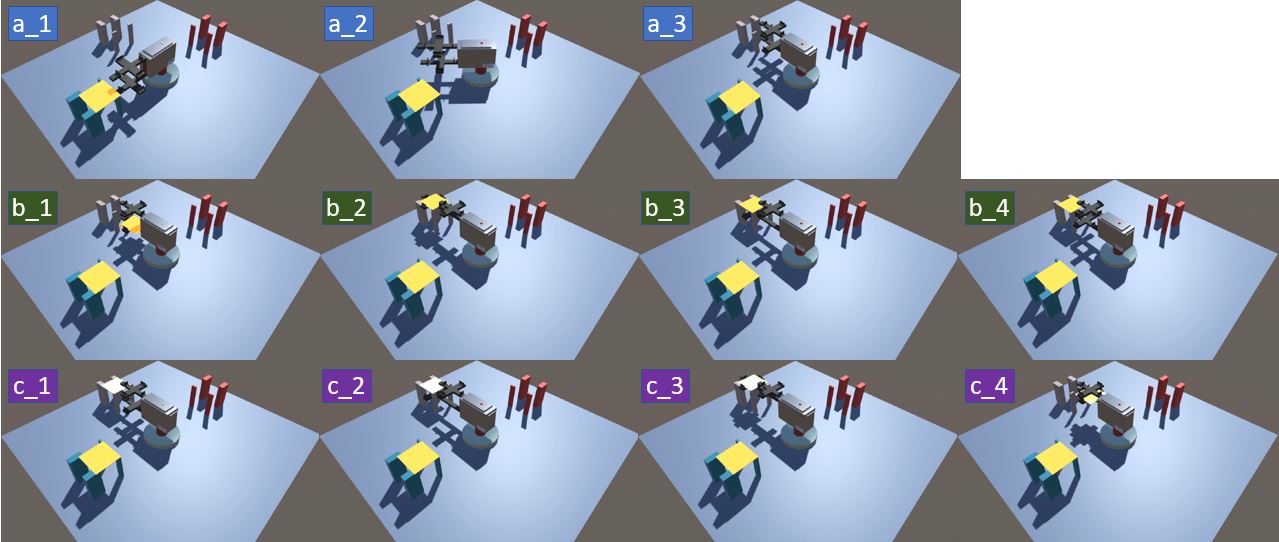}}}
    \caption{Actions in the simulation (a) robot rotation, (b) glass unloading, and (c) glass loading.}
    \label{fig:action}
\end{figure*}

To execute the training code based on the PPO algorithm, it is necessary to specify constants, such as the limit range $\epsilon$. In this study, the constants are designated as summarized in Table \ref{table:k}. The batch and buffer sizes are set to realize deeper learning, $\epsilon$ is the surrogate function value of the PPO algorithm, and the learning rate means the degree of update once. $\gamma$ refers to the rate of decrease of the previous reward mentioned in the previous value function.

\begin{table}[t]
\caption{Reward and penalty data (Basic experiment).}
\label{table:r}
\begin{center}
    \begin{tabular}{c c c}\hline
    Reward & Action & Value\\\hline
    Reward 1 & Processed glass arrives at the unloader  & +4 \\
    Penalty 1 & Time penalty   & -0.01 \\
    Penalty 2 & Glass dropped  & -1 \\
    Penalty 3 & Glass broken   & -1 \\
    Penalty 4 & Incomplete glass arrives at the unloader  & -1 
      \\\hline
    \end{tabular}
\end{center}
\end{table}

\subsubsection{Training through simulation}

Unity uses its own software as an environment for reinforcement learning and provides service ML agents that allow other learning routines to proceed based on the learning of neural networks in Python. In this manner, the environment based on the Python code can be tested in the Unity simulator. Fig. \ref{fig:position} and \ref{fig:distance} present the relevant observation schematics. It accepts the position data of the simulator and calculates and stores the position information of each chamber, current position of the robot and glass, and distance data between them in real time. The above data are then directly input into the neural network.

Fig. \ref{fig:action} shows the action schematic. The actions are loading and unloading depending on the rotation of the theta axis of the transfer robot, which is the agent, as well as the forward, backward, ascent, and descent motions of the robot arm.

\begin{figure}[t]
    \centering
    \framebox{\parbox{0.3\textwidth}{\includegraphics[width=0.3\textwidth]{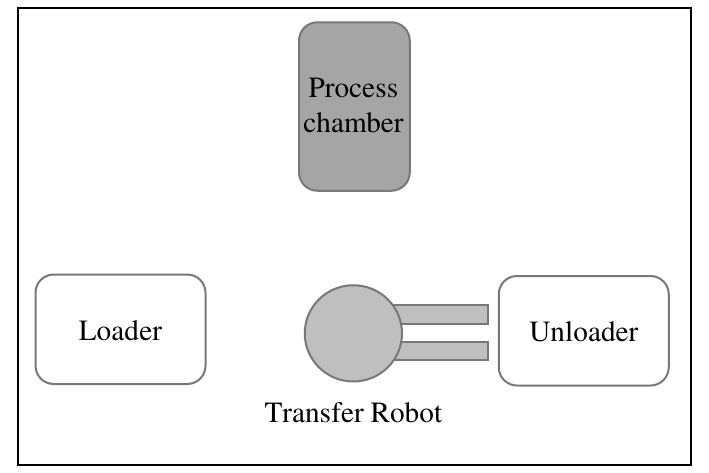}}}
    \caption{Extension experiment environment.}
    \label{fig:extensionenv}
\end{figure}

\subsection{Extension experiment}
The experimental configuration is identical to that of the basic experiment. In the extension experiment, a process chamber is added, as shown in Fig \ref{fig:simulation_cofig}, and a two-arm robot is used.

In a physical simulation, the state information changes continuously when an action is in progress. As illustrated in Figure 13, a reward is obtained when a reward condition occurs even as the action is in progress. Consequently, inaccurate experiences are accumulated in the replay buffer required for reinforcement learning. For this reason, we check the action that generates the reward and use the $s_{t+1}$ data directly for training. As the number of states increases, the training speed decreases. To prevent this, the state observation information is reduced. Instead of the existing position and distance information, only the glass state information of each chamber and the direction information of the robot are observed. 

As shown in the right-most of Fig. \ref{fig:position}, a glass state detection sensor is installed in each chamber. The hyperparameter settings are summarized in Table \ref{table:kk}. The batch size and buffer size are reduced according to the observed state information and the replay buffer. $\epsilon$ and $\gamma$ are increased and decreased, respectively, to satisfy the randomness of the action.

\begin{table}
\caption{Hyper parameters in PPO algorithm (Extension experiment.)}
\label{table:kk}
\begin{center}
    \begin{tabular}{c c || c c || c c}
    \hline
    Parameters & Value & Parameters & Value & Parameters & Value \\\hline
    Batch size & 32 & $\beta$ & 500 & Buffer size & 63 \\
    $\epsilon$ & 1.0 & $\gamma$ & 0.01 & $\lambda$ & 0.95 \\
    Learning rate & 3.0e-4 & Max steps & 5.0e7 & Memory size & 1536 \\  \hline
    \end{tabular}
\end{center}
\end{table}

\section{Experiment result}
In the same manner as in an actual FAB process, glass was continuously input, the process was performed normally by manually operating the transfer robot, and an experiment was conducted to move the finished glass to the unloader chamber. By adjusting the friction between the glass and the arm and using the guide pin of the arm, we were able to complete the process without glass separation.\\
\indent The experiment was conducted in three stages. First, process parameter optimization was evaluated in the implemented simulation environment. Second, a basic experiment was conducted under two chamber conditions by using a one-arm robot and the corresponding parameters. Finally, a three-chamber condition, that is, a confirmed evaluation, was performed using the two-arm robot.

\subsection{Process parameters split test}

\begin{figure*}[t]
    \centering
    \framebox{\parbox{1\textwidth}{\includegraphics[width=1\textwidth]{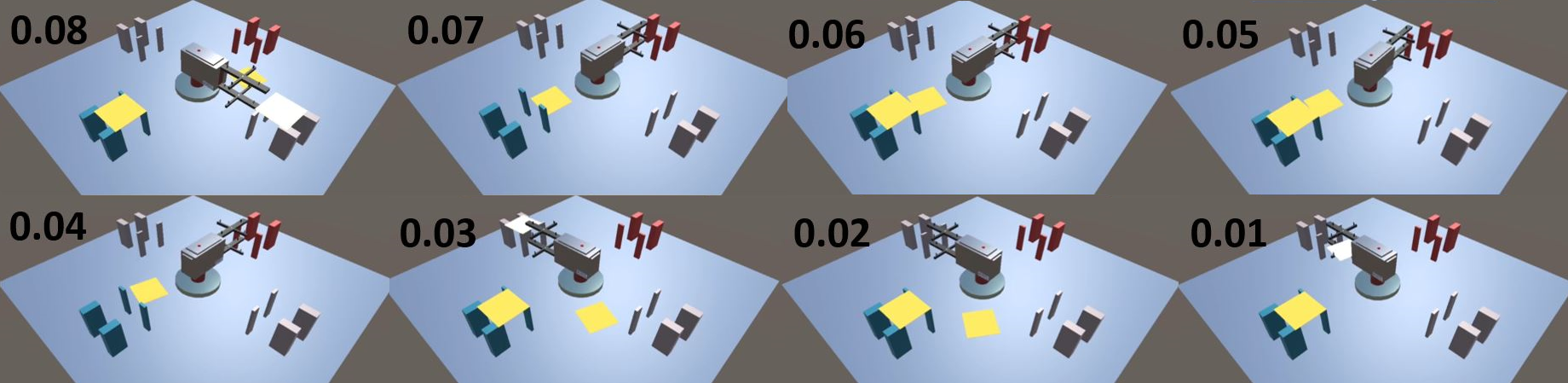}}}
    \caption{Rotation speed change test of transfer robot.}
    \label{fig:speedtest}
\end{figure*}

Various process parameter changes were evaluated in a simulation environment where hardware and software settings were completed. Various process parameter changes were evaluated in a simulation environment with complete hardware and software settings. By changing individual parameters, it was possible to obtain parameter values without glass loss. Fig \ref{fig:speedtest} shows the rotation speed change test of the transfer robot. The higher the speed, the greater was the amount of glass that could be moved, but at a rotation speed higher than 0.01, glass breakage occurred. In this manner, the required changes in the process parameters were evaluated, and the obtained results are summarized in Table \ref{table:parameter}.

\subsection{Basic experiment: One-arm robot + loader chamber + unloader chamber}
\begin{figure}[t]
    \centering
    \framebox{\parbox{0.5\textwidth}{\includegraphics[width=0.485\textwidth]{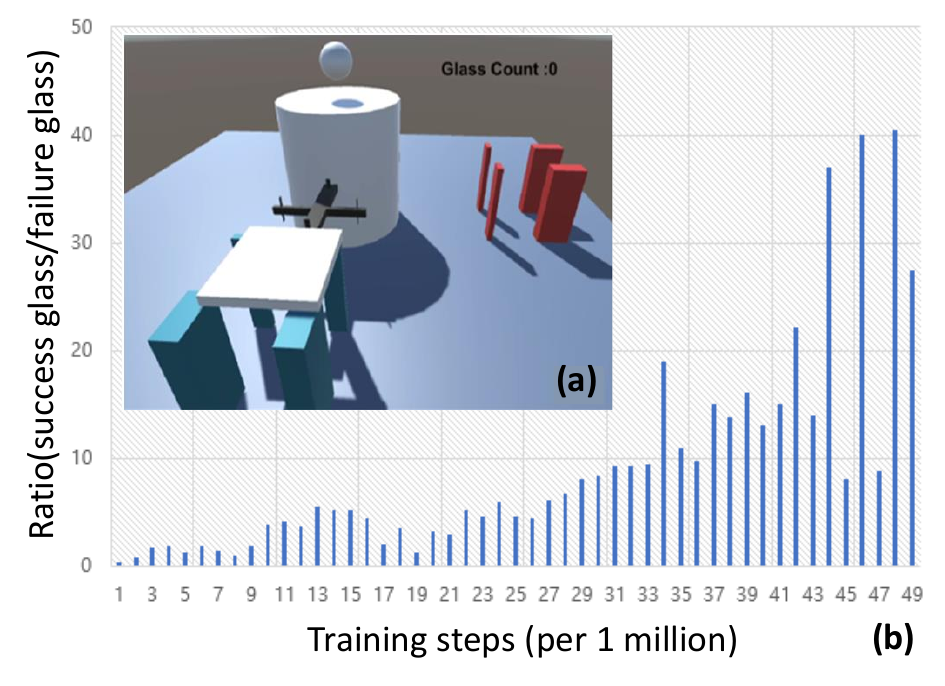}}}
    \caption{(a) Configuration of the simulation environment, and (b) training result.}
    \label{fig:test1result}
\end{figure}

A total of 10,409 glasses were input in the training phase. Among them, 2,165 glasses failed due to dropping and breakage, and 8,326 glasses arrived successfully at the unloader. In the training process, as shown in Fig. \ref{fig:test1result}, at the beginning, many movements and penalty actions were not related to the glass, but the number of successes and breakages increased and decreased, respectively, after 80,000 steps, and a maximum success ratio of 40:1 was achieved.

\subsection{Extension experiment: Two-arm robot + loader chamber + process chamber + unloader chamber}
In the basic experimental setting, in the simulation environment in which the number of robot arms was increased to two and the process chamber was added, the reward did not increase even when the learning progressed, and the success rate, too, did not increase. In the training process, the agent did not move the glass to the process chamber, and even if the movement was successful, it did not wait for process completion. Moreover, few attempts were made to move the finished glass to the unloader. To improve the learning accuracy, learning was performed using the state information of the moment at which the action accompanied by the reward occurred. To increase the learning speed, the state observation dimension was reduced.

\begin{table}[t]
\small
\caption{Process parameters.}
\label{table:parameter}
\begin{center}
    \resizebox{0.45\textwidth}{!}{
    \begin{tabular}{Sc Sc Sc}\hline
    Parameter & Initial value range & Optimal target value\\\hline
    Glass input interval & -  & 20s \\
    Glass size & 1500x1850mm(6G)   & 1000x1000mm \\
    Glass weight & 1$\sim$5kg  & 500 \\
    Number of chambers & 2$\sim$10ea   & 3ea \\
    Placement of chambers & -  & Cluster type \\
    Transfer speed & -  & 0.01 \\
    Transfer robot type & - & two-hands type \\
    Process time & - & 30s 
      \\\hline
    \end{tabular}
}
\end{center}
\end{table}

\begin{figure}[t]
    \centering
    \framebox{\parbox{0.5\textwidth}{\includegraphics[width=0.485\textwidth]{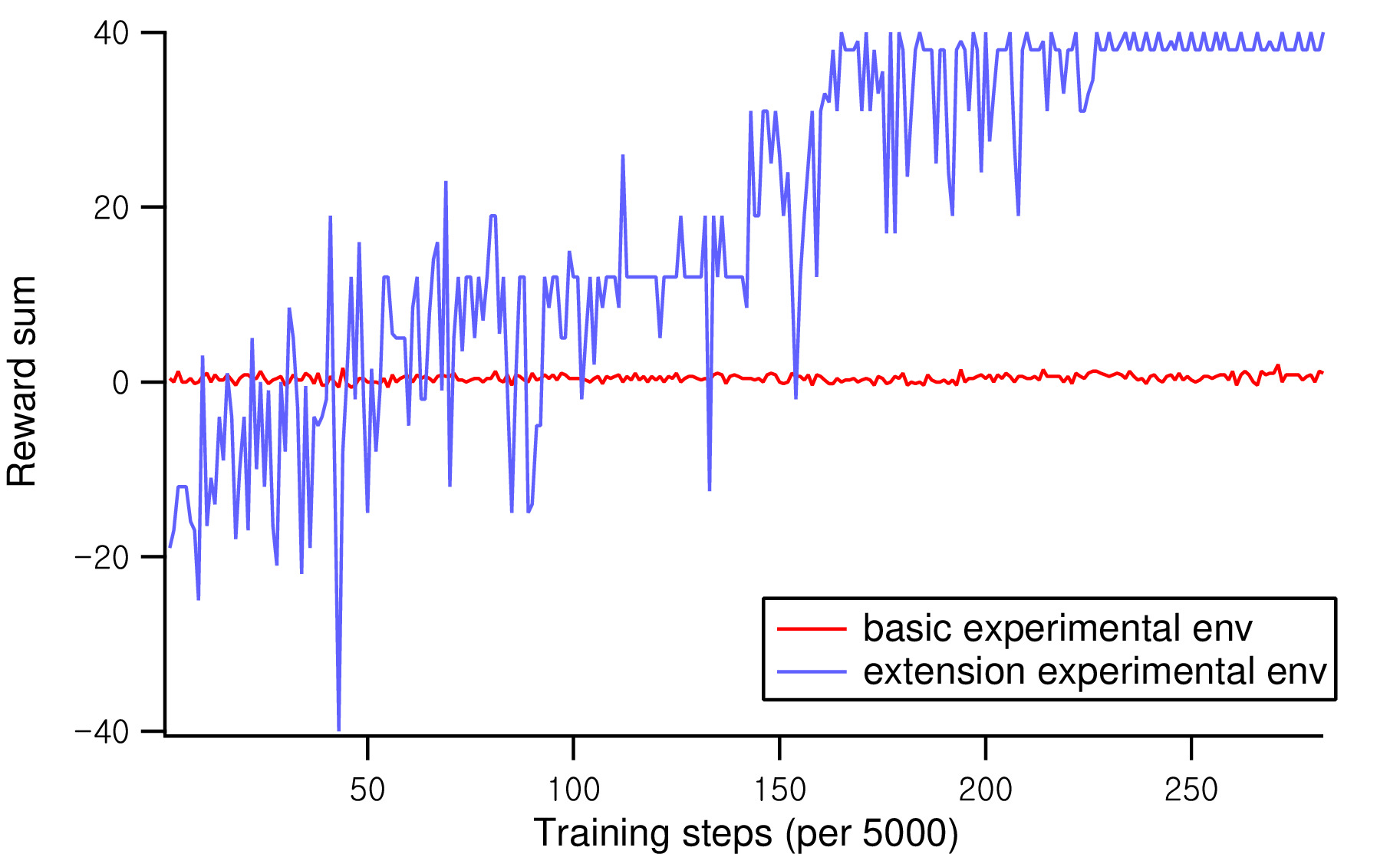}}}
    \caption{Two-arm robot, three-chamber layout: Reward sum according to test conditions.}
    \label{fig:result}
\end{figure}

After testing the expansion condition in the basic experimental environment, the compensation and convergence did not increase, as shown in Fig \ref{fig:result}, but in the expanded experimental environment, the maximum reward obtainable from a unit step was obtained. In the simulation action sequence, two glasses were loaded from the loader and moved to the process chamber, and the process-completed glass was simultaneously moved to the unloader chamber. 

\begin{figure}[t]
    \centering
    \framebox{\parbox{0.5\textwidth}{\includegraphics[width=0.485\textwidth]{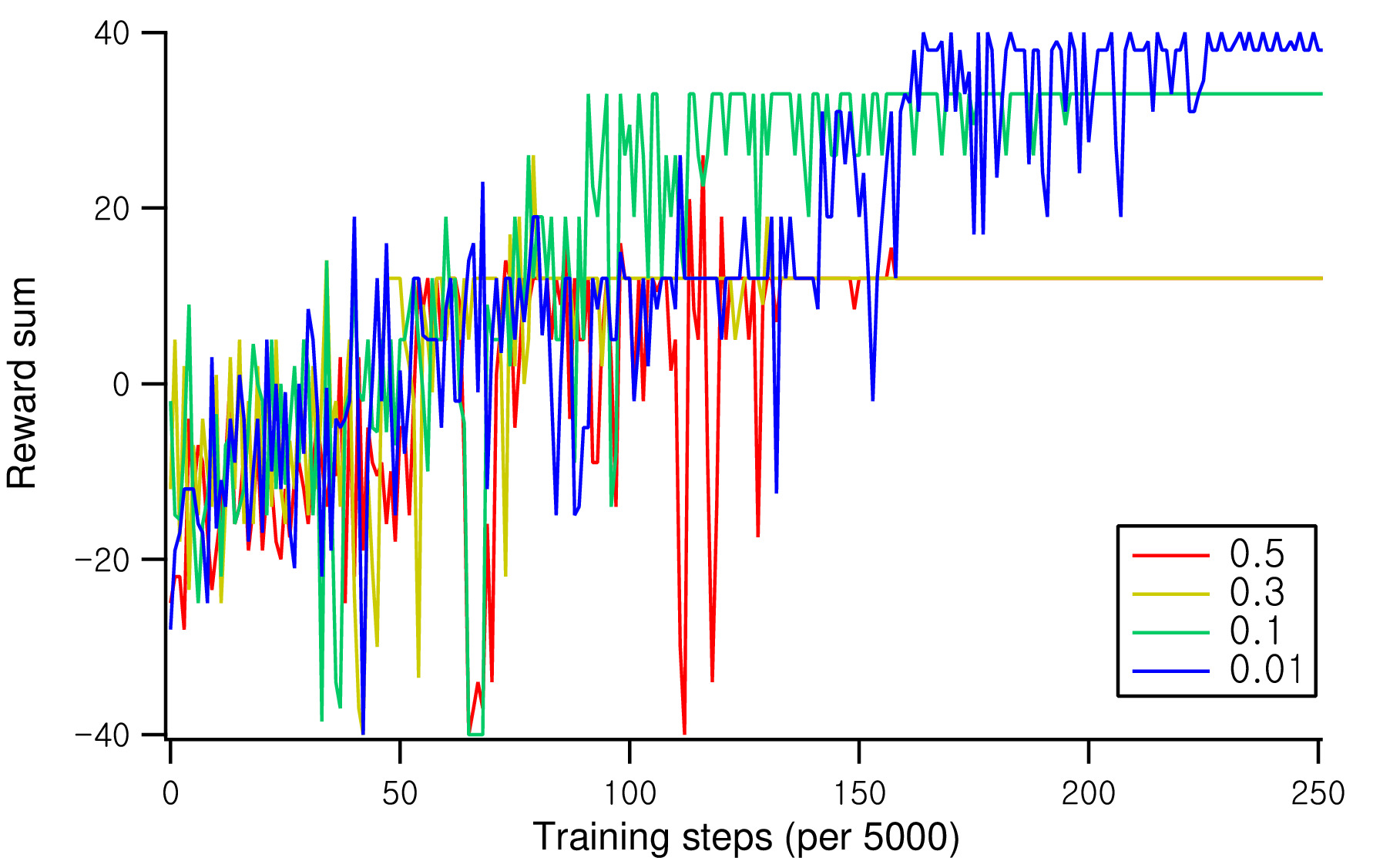}}}
    \caption{Split test of discount factor $\gamma$.}
    \label{fig:gammatest}
\end{figure}

This was the same as the order of glass processing in an actual FAB. A key hyperparameter of the scaling experiment was the discount factor $\gamma$. As shown in Fig \ref{fig:gammatest}, the smaller the gamma value, the more optimal is the scheduling. In this manner, it was possible to secure the randomness of learning by reducing the connectivity between the trajectories.

\section{Conclusion}

The flow control system is one of the most critical concepts in the display manufacturing industry. Optimal design of the flow control system that can fulfill diverse market demands is important to ensure that a FAB unit can maintain high productivity while manufacturing diverse products. A flow control system must be designed and operated considering this goal. In each stage of the system, the production rate must be increased, time must be saved, and product quality must be secured. From this viewpoint, simulations were performed to suggest an efficient design and operation plan for display manufacturing logistics management systems.

Manufacturing conditions similar to those in real FAB environments were implemented in the simulation with physical parameter information, and various process parameter split tests were performed in the simulation. It was configured to identify the cause of the problem in the simulation environment and solve the problem by changing the process conditions. Moreover, we demonstrated that the performance of the logistics management system was improved by using reinforcement learning. Based on this result, a FAB unit can predict, correct, and evaluate the improvement of serious problems. We expect that it will be possible to pre-evaluate the design of new production lines for future technologies, such as small-lot production and smart factories. In the future, we plan to test the learning of various machine learning algorithms in more diverse virtual display manufacturing environments.


\bibliographystyle{IEEEtran}
\bibliography{IEEEabrv, main}

\begin{IEEEbiographynophoto}{Hwajong Lee} received the B.S. degree in electronic and electrical engineering from the Sungkyunkwan University, Suwon, South Korea, and M.E. degree in engineering practice from the Seoul National University, Seoul, South Korea, in 2021. He is a senior engineer at Samsung display, South Korea. His current research interests include machine learning in manufacturing process.
\end{IEEEbiographynophoto}

\begin{IEEEbiographynophoto}{Chan Kim} received the B.S. degree in automotive engineering from the Hanyang University, Seoul, South Korea. He is currently pursuing the Ph.D. degree in electrical and computer engineering, Seoul National University. His current research interests include reinforcement learning, and autonomous driving.
\end{IEEEbiographynophoto}

\begin{IEEEbiographynophoto}{Seong-Woo Kim}(M'11) received the B.S. and M.S. degrees in electronics engineering from Korea University, Seoul, Korea, in 2005 and 2007, respectively, and the Ph.D. degree in electrical engineering and computer science from Seoul National University in 2011. He was a postdoctoral associate with the Singapore-MIT Alliance for Research and Technology. In 2014, he joined the Seoul National University, where he is currently an Associate Professor in the Graduate School of Engineering Practice. Dr. Kim won the first prize at Haedong paper award competition of the 21st International Conference on Electronics, Information, and Communication. He received the Best Student Paper Award at the 11th IEEE International Symposium on Consumer Electronics.
\end{IEEEbiographynophoto}

\end{document}